# Machine Learning for Structured Clinical Data

Brett Beaulieu-Jones[1]

[1] Institute of Biomedical Informatics, Perelman School of Medicine, University of Pennsylvania, D200 Richards Hall, 3700 Hamilton Walk Philadelphia PA, 19104.

Email: brettbe@med.upenn.edu

**Abstract**

Research is a tertiary priority in the EHR, where the priorities are patient care and billing. Because of this, the data is not standardized or formatted in a manner easily adapted to machine learning approaches. Data may be missing for a large variety of reasons ranging from individual input styles to differences in clinical decision making, for example, which lab tests to issue. Few patients are annotated at a research quality, limiting sample size and presenting a moving gold standard. Patient progression over time is key to understanding many diseases but many machine learning algorithms require a snapshot, at a single time point, to create a usable vector form. Furthermore, algorithms that produce black box results do not provide the interpretability required for clinical adoption. This chapter discusses these challenges and others in applying machine learning techniques to the structured EHR (i.e. Patient Demographics, Family History, Medication Information, Vital Signs, Laboratory Tests, Genetic Testing). It does not cover feature extraction from additional sources such as imaging data or free text patient notes but the approaches discussed can include features extracted from these sources.

## 1.1 Introduction

Precision medicine has the potential to substantially change the way patients are treated in many facets of health care. Precision medicine is the idea of delivering personalized treatment and prevention strategies by considering the holistic patient, including their genetics, environment, and lifestyle (Collins and Varmus 2010). Machine learning using structured clinical data will likely play a large role in the success or failure of precision medicine. Specifically, machine learning using structure data can help in finding associations between a patient's genotype and phenotype, identifying similar patients and predicting the efficacy of different clinical treatment strategies on a personalized level.

The amount of digital data collected in the clinic has rapidly expanded, the first EHRs are now more than 20 years old. The United States federal government mandated meaningful use of EHRs by 2014. According to the American Hospital Association, by 2015, 96% of acute care hospitals had implemented a certified EHR . Correspondingly, several top research institutions across the country have established departments or institutes in biomedical informatics using the EHR as a major data source in the past 5 years.

Smartphones, wearable devices and in-clinic diagnostic tools offer the ability to stream accurate measurements in real time. AliveCor received FDA approval in 2012 for its iPhone-based heart monitor using machine learning to detect Atrial Fibrillation in seconds. Billions of dollars in venture capital are currently being invested in companies, such as Grail, Foundation Medicine, and Guardant health, promising less invasive biopsies, or liquid biopsies, using machine learning to classify patients from circulating tumor cells in the bloodstream. Preventative wellness clinics, such as Forward, are emerging to characterize and track what it means to be healthy.

These are only a few examples of the many opportunities centered on patient data. Data for both evidence-based clinical decision making and computational research is becoming increasingly available and we must now develop new methods to preprocess and analyze this data at a matching rate.

## 1.2 Uses of Machine Learning for Structured Clinical Data

Each time a patient interacts with a health system, actions, notes, and measurements are recorded in the EHR. This wealth of data has made the EHR the primary source of structure clinical data. Three popular research applications of EHR data are:

>1.) Patient clustering to identify similar cases.
>2.) Electronic phenotyping for genetic studies.
>3.) Advising clinical treatment strategies.

These tasks can be performed using machine learning, but each task requires careful preprocessing of data and appropriate phrasing of the problem to utilize traditional machine learning methods. The nature of EHR data places emphasis on unsupervised clustering and semi-supervised classification. In this section, we discuss these common tasks and show examples where researchers have utilized machine learning effectively to guide discovery. There exist many great resources for understanding machine learning approaches as applied to general problems (Bishop 2006). We concentrate on how to position relevant clinical questions and the challenges specific to the EHR that need to be solved in order to apply these powerful techniques.

### 1.2.1. Patient/Disease Stratification

As we learn more about the mechanisms and etiology of a disease, our diagnoses can become more precise, leading to the creation of disease subtypes. Historically, cancers were diagnosed based on their occurrence location and their reaction to different treatments. As the mechanisms of cancer are better understood, they are further categorized by their physiological nature. The progression of subtypes in lung cancer illustrates the increases in resolution over time for a previously poorly defined disease (Kreybe L. 1962). Beginning with a single diagnosis based on occurrence in the lung, it was later differentiated as small cell lung cancer and non-small cell lung cancer (Mountain 1997; West et al. 2012). Non-small cell lung cancer was then broken up into squamous cell carcinoma, adenocarcinoma, and large cell carcinoma. Today these subtypes continue to be broken up based on the genetic locations and pathways of associated risk variants.

What happens when physiological differences cannot easily be used to subtype disease? This is true with several metabolic disorders, for example, metabolic syndrome has been redefined numerous times. It is associated with a wide range of comorbidities and presents in a clinically heterogeneous manner. These comorbidities, including coronary heart disease, diabetes, and stroke, represent an oversized risk to public health and increasingly unwieldy burden on the health care system. Despite this, metabolic syndrome's predictive value for cardiovascular events, disease prediction and progression is disputed and may not outperform the individual components it's made up of (Shin et al. 2013). While the concept of identifying patients at high risk of developing diseases such as heart disease and diabetes for early intervention is an important one, metabolic syndrome in its current form fails to do this effectively.

Li et al. demonstrated the ability to identify disease subtypes of patients with a metabolic disorder, type 2 diabetes (Li et al. 2015). To do this they performed a topological analysis of 11,210 patients with type 2 diabetes at Mount Sinai Medical Center in New York. This topological analysis constructed a network of patients by connecting those most similar to each other. Using this they found three unique subtypes. Subtype 1 demonstrated the traditional observations of type 2 diabetes, hyperglycemia, obesity, and eye and kidney diseases. Subtype 2's main comorbidity was cancer, and subtype 3's unique comorbidities were neurological diseases. These subtypes are likely enriched for etiological differences; the disease likely operates differently in someone who develops cancer than someone who develops kidney disease. By developing a machine learning classifier to identify which subtype a patient is in as early as possible, clinicians may be able personalize treatment to reduce the odds of developing these more serious comorbidities.

Multiple sclerosis (MS) illustrates an additional area machine learning for disease stratification could be particularly useful. Multiple sclerosis was traditionally subtyped into Relapsing-Remitting MS and Progressive MS. In 2014, it was recommended that these subtypes be further divided into six total subtypes (Lublin et al. 2014). Unfortunately, the current strategies for determining subtype and thus treatment strategy require looking at the progression of the disease. This is essentially a retrospective diagnosis and means personalized treatment plans cannot be started until progression has been observed. Could unsupervised clustering be used to identify subtypes earlier on?

### 1.2.2. Electronic phenotypes for Genetic Associations

Genetic associations examine whether a genetic variant is associated with a specific trait (Figure 1). This specific trait, a phenotype, can be a moving target when dealing with the complexity of human disease. The trait is often a human defined disease. Those with the disease are labeled the case and those without the disease are considered controls. Early genetic associations using the electronic health record were performed with raw International Classification of Diseases (ICD) codes. ICD codes are recorded by physicians when diagnosing a patient with a condition, and are used to ensure proper billing and insurance reimbursement. ICD codes are published and updated by the World Health Organization and are primarily used for clinical billing purposes. Despite ICD-10 being initially published in 1994, ICD-9 codes are still commonly used in both clinical and research settings.

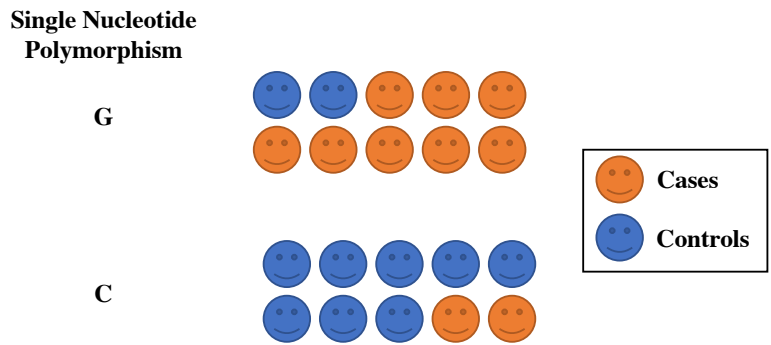

**Figure 1**. In this example, the guanine (G) nucleotide is more common in the cases than the cytosine nucleotide.

While ICD codes provide a clear, discrete endpoint for genetic associations, the use of billing codes can introduce unintentional biases to analyses. An ICD code may be added to an EHR in order to issue and receive insurance reimbursement for a test to screen for the disease the ICD code represents. In this case, not only is the timing of diagnosis difficult to determine, but solely looking at the ICD codes for a patient is likely to introduce false positives. In addition, certain ICD codes are more easily reimbursed than others. When a clinician determines that a patient requires a treatment or test to increase their odds of a successful outcome, the clinician is incentivized to choose the ICD code most likely to allow them to effectively treat their patient.

Phenotype algorithms can be developed using the structured EHR to leverage both ICD codes and the rest of the of a patient's record. The eMERGE project is a national network which has deployed phenotype algorithms for over 40 diseases, over 500,000 EHRs and 55,000 patients with genetic data. Many of the phenotype algorithms are simple rules-based systems, for example: Type 2 Diabetes (Figure 2).

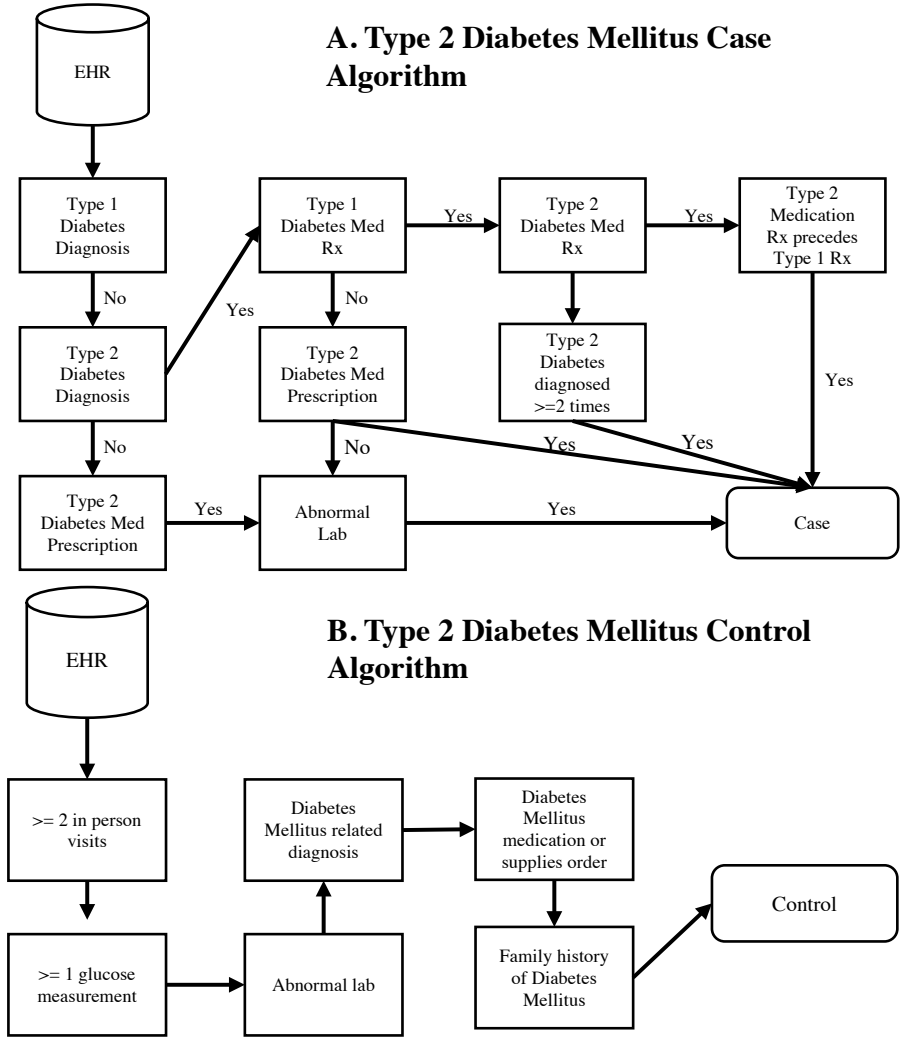

**Figure 2**. Phenotype Algorithms for Type 2 Diabetes Mellitus. **A.)** Case selection from the EHR. **B.)** Control Selection from the EHR. Adapted from: (Khardori 2014)

An approach to study phenotype-genotype associations from the EHR are Phenome-wide association studies or PheWAS (Denny et al. 2010). PheWAS use EHRs to define a phenome that can be linked back to individual genetic variants. The approach can discover gene-disease associations while identifying pleiotropic effects of individual SNPs. PheWAS generally uses the ICD9 codes to construct a phenotype. While primarily used for billing, these codes provide a set of discrete variables that can represent many phenotypes for a patient at the same time, providing greater resolution. Besides the repurposing of billing codes, a major challenge of PheWAS is in understanding the functional mechanisms at work behind GWAS SNP matches. Stratification by the 4,841 different codes creates wide data, presenting statistical challenges in achieving adequate power. This challenge of achieving adequate power will be exacerbated by the transition to ICD10, with less historical data built up and the potential for over 16,000 codes. Continuing to increase open data access will allow researchers to utilize a more accurate phenotypic representation while lessening the burden of statistical challenges. Coding systems, unlike patient notes or genomic data should be easier to anonymize, aggregate, and distribute.

Because ICD billing codes can be biased, as evidenced by phenotype algorithms having multiple steps to catch errors for both case and control status, using billing codes alone may cause misclassification of phenotypes. The misclassification of phenotypes substantially reduces the power to detect linkage in case-control studies. With 1% phenotypic misclassification up to 10% of the power is lost, and with 5% phenotypic misclassification, the power is reduced by approximately two-thirds (Gordon et al. 2004; Buyske et al. 2009; Manchia et al. 2013). Misclassification can occur for a variety of reasons including misdiagnosis/clinical error, clerical error, or lack of scientific knowledge about the disease in question.

Labbe et al. showed increased linkage by clustering lifetime symptoms in schizophrenia and bipolar disease to form more homogenous phenotypes. Separating cases by the symptoms of psychiatric diseases compensates for the inability to subtype these diseases by physical properties (Labbe et al. 2012). This is important due to the deficit of physiological understanding for these diseases. Labbe et al. also included familial information to understand the heritability of these diseases. When looking at subtypes that show a strong familial aggregation they observed higher linkage scores. By looking at ancestral histories for subtypes, the expected heritability could be better estimated resulting in a reduction of "missing heritability."

Phenotypic subtyping was also used successfully in the analysis of genetic variants responsible for the severe development regression and stereotypical hand movements of Rett syndrome. Causal mutations were found in the FOXG1 and MECP2 genes and deletions at the 22q11.2 locus (Chaste et al. 2015).

Each of these examples point towards the promise of using machine learning to cluster patients based on their EHRs to identify disease subtypes or more homogenous groups of patients for use in association studies.

### 1.2.3. Clinical Recommendations

The availability of data and advances in biomedical informatics have helped to make medicine increasingly evidence based and in some cases entirely data driven. Clinicians and researchers now have the ability to leverage millions of data points when designing and determining treatment best practices. The New England Journal of Medicine recently held the SPRINT data analysis to "use the data underlying a recent article to identify a novel clinical finding that advances medical science." The original clinical trial sought to see whether intensive management of systolic blood pressure (<120 mm Hg) was more effective than standard management (<140 mm Hg). The original trial was stopped early due to the success of the intensive management strategy in reducing cardiovascular events. The data from the trial was released as a challenge where teams used machine learning approaches (primarily rules based) to provide personalized recommendations.

More personalized treatment strategies are a popular use of machine learning in the EHR. This can be driven by genomics (pharmacogenomics), or simply by sub setting patients based of attributes (race, BMI, etc.). Wiley et al. demonstrate the importance of training an algorithm on a population similar to the application population (Wiley et al. 2016). In their case it was necessary to extract the perecent African ancestry from the genome instead of self reported race in order to improve the model fit.

Due to the inherent risk of adjusting clinical treatment strategies, many of the early applications of machine learning in health systems have been seen in academic research (retrospective analysis, drug development, pharmacogenomics) and for things like resource usage. For example, how likely is a patient coming into the ER to need an ICU bed? Increasingly machine learning methods are likely to be applied to clinical decisions including providing prognosis information for shared decision making strategies. Deep learning, in particular, is becoming an increasingly tool for drug discovery and development (Ching 2017).

## 1.3 Challenges of using Machine Learning in the Structured EHR

### 1.3.1. Limited "gold-standards"

Large institutions and health care systems can have EHRs containing millions of patients and billions of measurements. Despite the size of these data, electronic phenotyping requires a gold standard to validate accuracy. This gold standard often requires time consuming, manual clinician review and is thus expensive.

In addition, the selection of cases and controls can unintentionally create biases in downstream algorithms and analyses. It is often easiest to select the most severe cases and the healthiest controls. In these circumstances researchers can have the greatest confidence they are accurately selecting a true case or control. Unfortunately, this creates a biased training set where it is difficult to differentiate between less severe cases and less healthy controls. Figure 3 shows an illustration of a simulated dataset where the first two principal components happen to represent the degree of the case phenotype. If the most severe cases are selected, a classifier trained to distinguish between cases and controls is unlikely to generalize well. If less severe cases are chosen, there may be issues with mislabeled cases. An example of another bias can be seen in the Type 2 Diabetes Mellitus algorithm, controls must have at least 1 glucose measurement (Figure 2B). For a young patient, this means that a clinician must have had reason to suspect that the patient's glucose could be abnormal and thus could bias controls to patients who "look" like they are at a high risk for developing Type 2 Diabetes.

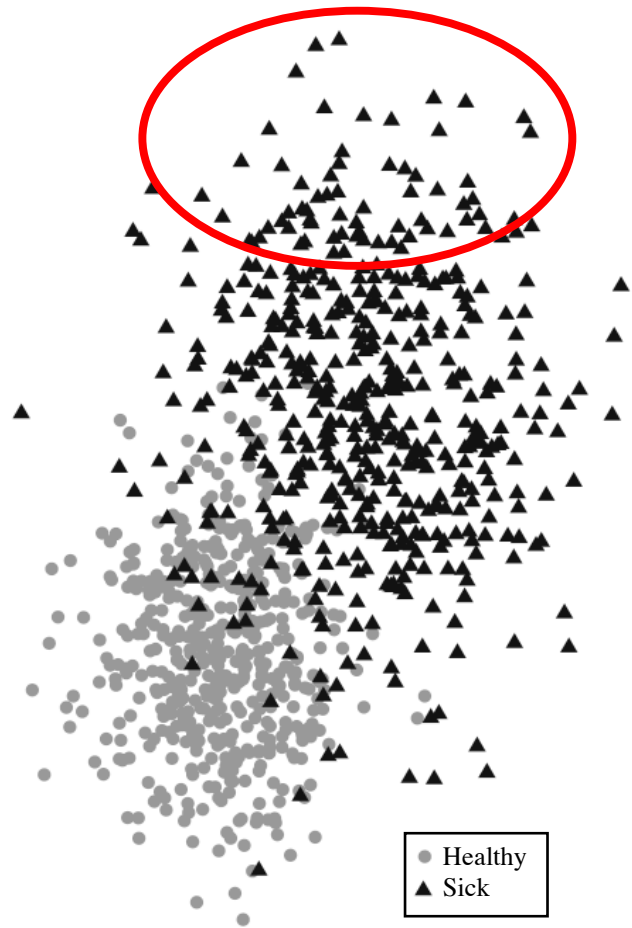

**Figure 3**. Simulated disease severity plot where the 2 principal components stratify patients according to severity.

Because patients move between health systems, a patient may not be diagnosed in the system they are treated in and may only have a partial history. Some methods for controlling for incomplete histories can result in smaller sample sizes. It is common to include only patients who have a visit in the system prior to the diagnosis of the phenotype of interest. While this can help to determine the diagnosis date for a disease, it excludes anyone who was diagnosed on their first visit to a particular health system.

### 1.3.2. Missing Data

The average patient is unlikely to have measurements for the clear majority of fields in the EHR. It does not make sense logistically or economically to administer every test to a seemingly healthy patient. There are three primary types of missing data:

1.) Missing Completely at Random (MCAR) – when data is missing in a completely unrelated way to the values of both the observed and the unobserved data.
2.) Missing at Random (MAR) – when the data is missing based on the observed data, when other fields in the EHR indicate whether the value will be present or absent.
3.) Missing Not at Random (MNAR) – when the data is missing based on the values of the unobserved data

Figure 4 shows the ability to use a random forest to predict whether a lab value will be present or absent based on other lab values. Unsurprisingly data that is MCAR cannot be predicted (Figure 4A), and data that is either MAR or MNAR (Figure 4B, 4C) can be predicted with an accuracy significantly greater than random. In practice, most missing data in the EHR tends to be of the MAR variety. Clinicians must decide which measurements are relevant and fiscally responsible, irrelevant tests are wasteful and it does not make sense to subject patients to unnecessary discomfort. The clinician is making these decisions based on the observations they make, so when data is missing it is related to the observed data.

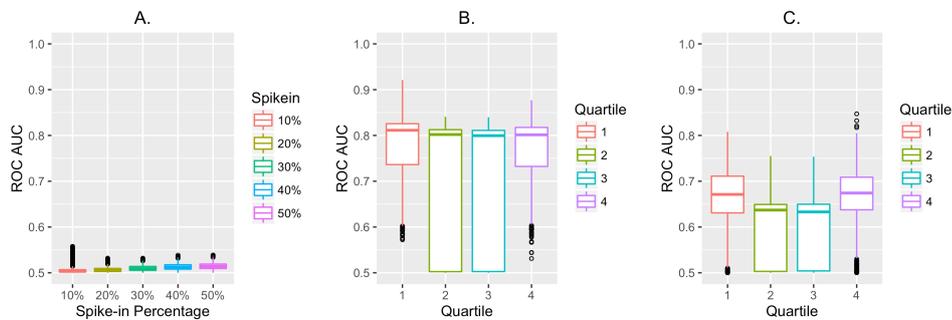

**Figure 4. A.)** Data that is missing completely at random cannot be predicted. **B & C.)** Data that is missing at random and missing not at random can be predicted. (To appear)

MCAR data is less likely to present issues to downstream analyses than data that is either MAR or MNAR, but if not handled correctly all three types of missing data can introduce unintentional biases to all sorts of downstream analyses including machine learning. Machine learning algorithms often expect a complete matrix as input and are not designed to handle null values. This often leads to researchers performing one of three options:

1.) Perform feature selection of relevant features and use only complete cases, or patients that have values for all features.
2.) Modify the algorithm to accept null inputs (often by ignoring them) or
3.) Perform imputation to predict what the value for a feature would be.

Each of these options have several pros and cons and can have unintended effects on machine learning. When performing complete case analysis after feature selection, the features included can lead to including either more severe cases or cases that were harder to diagnose. Imagine a disease that is diagnosed by a laboratory measurement where values over 10 conclusively indicate you have the disease but values between 8 and 10 require an additional test. If the additional test is included in the features selected, the complete cases are now only the patients that were harder to diagnose. When modifying an algorithm to accept null inputs, the researcher needs to be careful that the algorithm does not disproportionately learn to depend on patients that have all of the measurements or only the most complete measurements. If the algorithm relies on patients with all of the measurements, many of the same issues that arise in complete case analysis repeat. If the algorithm learns to ignore rare measurements it can miss signal. For example, in an analysis of different treatment options, say 40 of 10,000 patients suffered a fairly rare but severe adverse event. Without careful monitoring the algorithm may not place enough importance on this feature despite the fact this outcome is disproportionately important.

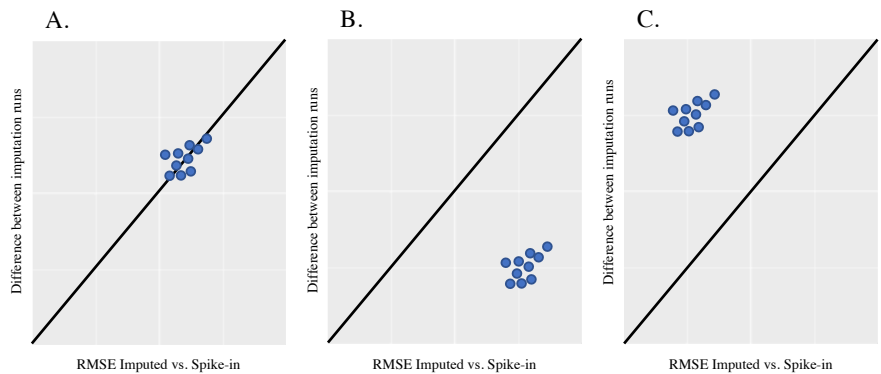

**Figure 5**. Comparison between spike in accuracy and variation between imputation runs.

Imputation can be effective in the EHR because many missing values can be inferred by omission and just knowing whether a value was present or absent can be useful. If a patient has never had a chest x-ray, it is unlikely that their physician suspects a broken rib cage. This information can be provided to downstream machine learning algorithms by performing imputation. It is, however, very important to carefully analyze the results of imputation. Oftentimes much can be learned simply by looking at which methods are the most accurate. Direct accuracy can be measured by spiking in missing values to replace known values, imputing these spiked-in values and measuring their difference from the real values. Despite this direct accuracy should only be used as a benchmark, and it is important to analyze the effect of imputation on the downstream analyses you are performing (Figure 5).

For example, mean imputation may perform strongly in an analysis using spiked-in missingness but remove all variance from the imputed values for a feature (Figure 5C). Other evaluation criteria, such as, comparing the variance between imputed values of different imputation runs and the difference between imputed values and real values. Ideally these values would be highly correlated in order to maintain the variance structure. Popular imputation methods for EHRs include K-Nearest Neighbors, Singular Value Decomposition and Multiple Imputation by Chained Equations.

All three of these strategies for handling missing data may introduce bias when performing EHR-based analyses. It is important to consider potential effects and ideally to utilize multiple strategies and examine the differences.

### 1.3.3. Privacy, Reproducibility and Data Sharing

Patient privacy needs to be a focus of any secondary use of EHRs. Because a patients EHR is 'de-identified' does not mean that it is anonymous. Latanya Sweeney demonstrated this emphatically when the Massachusetts Group Insurance Commission released de-identified data on state employees (Shaw 2009). These records included each hospital visit and Sweeney was able to re-identify several patients including the former Governor of Massachusetts. Sweeney was able to do this from his birth date, zip code and sex alone, and to prove a point mailed the Governor a copy of his personal records. The task of re-identification has been shown possible in several other cases where data holders attempted to share their data, including the Netflix challenge. Narayana and Shmatikov were able to de-anonymize users in the Netflix challenge by linking their viewing histories with popular movie review sites. For users who had rated more than 6 movies, they were able to do this with greater than 90% accuracy (Narayanan and Shmatikov 2008). This, in part, led to Netflix canceling the second iteration of its popular recommendation contest following a privacy lawsuit.

Caution needs to be taken even when the actual data is not released. Deep learning models can have many millions of parameters, allowing adversaries to perform membership inference attacks in order to determine whether a user was a member of the study or not. Shokri et al. (Shokri et al. 2016)) demonstrate greater than 90% precision even with 30,000 examples in the training set. They do this by examining the trained parameters of a deep neural network trained on the CIFAR-100 dataset.

Even without the model, enterprising adversaries performed membership inference attacks with only black-box access to the target model through an API. Shokri et al. again demonstrated this on various purchase history datasets made available through Amazon and Google APIs.

One approach to adding privacy protection is called "Differential Privacy" (McSherry and Talwar 2007). Differential Privacy is a robust, meaningful and mathematical rigorous definition of privacy which operates under the knowledge that data cannot be fully anonymized and remain useful. If you remove all of the signal in a dataset to anonymize, machine learning methods are fruitless. If you keep any signal at all, there is a chance an adversary will able to discover information about the members of the dataset. The goal of differential privacy is to find a balance between an acceptable risk, the privacy budget, and usefulness of the data. It attempts to minimize the likelihood an adversary can perform a membership inference attack to determine if a subject is in a dataset. It works by adding a plausible deniability of any outcome by inserting random noise into the information made available. If balanced, meaningful answers can be interrogated from the data while greatly reducing the risk that any member of a study is harmed by de-identification. A classic example and simple way to think about differential privacy is to imagine a study where participants are told to answer a question. Before answering the question they flip a coin, if the coin lands heads, they give the real answer, the truth. If the coin lands tails, they answer randomly by flipping an additional coin and responding yes if it lands heads and no if it lands tails.

Simmons et al. used a variant of differential privacy to enable privacy preserving genome wide association studies even when there is significant population stratification. Genomic data has a high dimensionality and relatively low signal to noise ratio making de-identification or other attempts at masking individual records impractical. They demonstrate the ability to allow users to query summary statistics while minimizing privacy risks. This is a particularly interesting application because while genome sequencing prices have rapidly decreased, the combined costs of recruitment and sequencing are a major barrier to this type of research.

Beaulieu-Jones et al. recently showed the ability to train deep neural networks under differential privacy to generate synthetic data that closely resembles the SPRINT clinical trial data (Beaulieu-Jones et al. 2017). This method allows for increased sharing of valuable, difficult to obtain datasets. Differential privacy is a rapidly growing area, we suggest "The Algorithmic Foundations of Differential Privacy" (Dwork and Roth 2013) as a starting point if interested in implementing differential privacy.

Privacy challenges can make sharing data prohibitively difficult. This in turn presents challenges in reproducing work from other researchers. Even if source code is shared, researchers attempting to reproduce original research generally can only compare final results. This means that even if a protocol of a paper is well written and described, if it has 100 steps, a researcher attempting to reproduce cannot be sure where their results diverged. Because of this challenge we strongly advocate publishing intermediate results. This can help narrow down divergences to a few steps, was it the data? The preprocessing? The actual analysis? The plotting into charts? One way to release intermediate results without adding a large amount of additional work is to use continuous integration to run the analysis and export the log file (Beaulieu-Jones and Greene 2017).

### 1.3.4. Longitudinal Data

A key attribute and potential strength of EHRs is the ability to track the way a patient progresses over time. Early moving caregivers such as Geisinger Health System implemented initial EHRs over twenty years ago but fully utilizing this longitudinal presents challenges to researchers.

Longitudinal EHR data are often irregular time series. Measurements are recorded at irregular times, can be mixed type (continuous, ordinal, categorical), require feature extraction (images, free text). It is common for researchers to take a single time point (i.e. current time, set time after diagnosis etc.) and use this as the single end point or label for machine learning analyses. This can be problematic when patients have arrived at that point through very different routes. For example, if using a systolic blood pressure as an end point, one patient may be on an intensive blood pressure management protocol while another with the same blood pressure may have never taken medication. In the SPRINT clinical trial there were patients on as many as seven medications to manage blood pressure (Group 2015), if unmedicated these patients would almost definitely have significantly higher measurements. One method researchers use to remediate this issue is to derive statistics to represent the time series, such as taking the median value. This can be insufficient when the way clinicians choose to observe and treat patients based on data either not recorded in the electronic health record, in the unstructured data or in fields not selected for inclusion can also bias the labels. For example, if patient A has a single normal white blood cell count, and patient B has had a monthly count every month for the past 5 years. A clinician could have been checking to see if patient A showed an increased white blood cell count after a surgery suspecting a possible infection. In contrast, the repeated measurements for patient B indicate the clinician may have a reason to believe patient B is immunocompromised or may become immunocompromised due to a virus or adverse reaction to a medication and is using the white blood cell count to monitor this. Despite the patients having relatively equal white blood cell counts, using this single value as a label is clearly inadequate to represent the complete state of the patient. For this specific case deriving a panel of statistics including features such as the count and variance of the measurement could help to better represent the current state of a patient. Recent work takes this further to calculate disease and patient trajectories by generating networks of the way a patient or disease progresses over time. Jensen et al. demonstrated this using 6.2 million patients from Danish National Patient Registry to cluster patients based on time dependent disease diagnoses (Jensen et al. 2014). These disease diagnoses were extracted from patterns of ICD-9 codes on patient's EHRs. This method creates a visualization of patient trajectories and allows for analyses of co-morbidities observed in health systems in order to identify important patterns that indicate the potential for more severe outcomes. Further work in this field could move beyond billing codes in order to allow for increased resolution of patient trajectories.

## 1.4 Future and evolving opportunities for Machine Learning in the Structured EHR

### 1.4.1. Quantitative Electronic Phenotyping

Traditionally, genetic association studies relied on binary outcomes as target phenotypes for the association. Quantitative trait loci studies provide the ability to measure correlation between DNA variation and a phenotype. Quantitative traits occur on a continuum and are driven by multiple genes in conjunction with the environment. By using clustering and other machine learning techniques, researchers can represent disease as quantitative rather than binary values. This has several advantages. Patients with a common disease that present with different symptoms, different levels of effect or different paths of progression can be clustered into homogenous subgroups with similar patients. These clusters are likely enriched etiologically, meaning the reasons the disease is causing each cluster are different. Within each cluster or across the entire spectrum of diseases, a phenotype can be constructed to better represent how severe of a case a patient has. Diseases where using a binary case control status has been effective are likely etiological homogenous, or so disruptive to a particular system that the severity is irrelevant. These represent the low hanging fruit, but many diseases present in heterogeneous manners (Cancers, Amyotrophic Lateral Sclerosis, Multiple Sclerosis, Alzheimers etc.). Fine-tuned quantitative phenotyping could have the ability to resolve homogenous subgroups, greatly increasing statistical power and creating a better target for association.

### 1.4.2. Deep Learning, Unsupervised and Semi-Supervised Learning Approaches

Deep Learning has already led to state of the art results in a variety of fields including image processing, speech recognition, and gameplay. Many of the early "wins" using deep learning and more generally machine learning in the EHR involve applications of algorithms proven successful in other domains. This has been particularly true in unstructured EHR data such as images (cancer tumor detection etc.) and natural language processing for free text. This is sufficient when EHR learning tasks resemble tasks popular among general machine learning researchers but long term advances will require specialized algorithms customized to the unique challenges presented by EHR-based research. Algorithmic development is only one part of the equation. The proper phrasing of problems and preprocessing of data will likely have as much if not more importance than algorithmic development.

Three of the pioneers of deep learning, Yann Lecun, Yoshua Bengio and Geoffrey Hinton wrote, "Unsupervised learning had a catalytic effect in reviving interest in deep learning, but has since been overshadowed by the successes of purely supervised learning… we expect unsupervised learning to become far more important in the longer term" (LeCun et al. 2015). The challenge of collecting labeled data for supervised learning in the EHR may be an ideal environment for a reemergence of unsupervised and semi-supervised learning approaches. Early examples show that deep autoencoders are adept at this task (Beaulieu-Jones and Greene 2016; Miotto et al. 2016).

### 1.4.3. Interpretation and the "Right to explanation"

For some clinical decisions, a black box algorithm with high accuracy is sufficient to improve medical care. An algorithm that can more accurately identify a tumor in imaging than a human has obvious benefits. For other problems, a black box is insufficient, it is unlikely to help researchers understand the physiology or etiology of a disease. In this setting, outside of a clinical decision, a less accurate but interpretable algorithm may be preferred. In addition, clinicians are likely to be skeptical and slower to adopt algorithms whose decisions cannot be rationally explained.

Furthermore, in April 2016, the European Union passed a data protection law entitled the "General Data Protection Regulation (GDPR) which will begin in 2018. The GDPR provides stricter conditions for sensitive data collection and storage, including, for example genetic and biometric data. It also sets regulation on privacy policies and further formalizes the "right to be forgotten." Of particular interest to the machine learning community is the language prohibiting decisions "based solely on automated processing and which produces adverse legal effects concerning, or significantly affects, him or her" and provides the right "to obtain an explanation of the decision reached after such assessment or to challenge the decision." It remains to be seen how this would be applied and if it will have any effect on the usage of artificial intelligence in the clinic but it demonstrates the fact that people want to understand how and why a decision affecting their wellbeing is made. Work to create high performing algorithms that provide interpretable, explainable decisions is increasingly important as clinicians increasingly rely on the aid of artificial intelligence.